\documentclass[11pt,a4paper]{article}
\usepackage[hyperref]{naaclhlt2019}
\usepackage{times}
\usepackage{latexsym}
\usepackage{amssymb}

\usepackage{tabularx}
\usepackage{pifont}

\usepackage{url}
\usepackage{graphicx}

\usepackage[]{algorithm2e}

\usepackage{multirow}
\usepackage{booktabs}

\usepackage{microtype}

\usepackage{gb4e}

\aclfinalcopy

\title{MCScript2.0: A Machine Comprehension Corpus Focused on Script Events and Participants}

\author{Simon Ostermann$^1$~~~ Michael Roth$^{1,2}$ ~~~ Manfred Pinkal$^1$ \\\ \\
  $^1$Saarland University~~~ $^2$Stuttgart University\\
  {\tt \{simono|pinkal\}@coli.uni-saarland.de  rothml@ims.uni-stuttgart.de} 
}

\date{}

\begin{document}
\maketitle
\begin{abstract}
  We introduce \textit{MCScript2.0}, a \textbf{m}achine \textbf{c}omprehension corpus for the end-to-end evaluation of \textbf{script} knowledge. MCScript2.0 contains approx. 20,000 questions on approx. 3,500 texts, crowdsourced based on a new collection process that results in challenging questions. Half of the questions cannot be answered from the reading texts, but require the use of commonsense and, in particular, script knowledge. We give a thorough analysis of our corpus and show that while the task is not challenging to humans, existing machine comprehension models fail to perform well on the data, even if they make use of a commonsense knowledge base. The dataset is available at \url{http://www.sfb1102.uni-saarland.de/?page_id=2582}
\end{abstract}

\section{Introduction}
People have access to a wide range of commonsense knowledge that is naturally acquired during their lifetime. They make frequent use of such knowledge while speaking to each other, which makes communication highly efficient. The conversation in Example \ref{ex:conversation} illustrates this: For Max, it is natural to assume that Rachel paid during her restaurant visit, even if this fact was not mentioned by Rachel.

\begin{exe}
	\ex Rachel: ``Yesterday, I went to this new Argentinian restaurant to have dinner. I enjoyed a tasty steak.''\\
	Max: ``What did you pay?'' 
	\label{ex:conversation}
\end{exe}

\textit{Script knowledge} is one of the most important types of commonsense knowledge and subsumes such information \cite{schank77}. It is defined as knowledge about everyday situations, also referred to as \textit{scenarios}. It subsumes information about the actions that take place during such situations, and their typical temporal order, referred to as \textit{events}, as well as the persons and objects that typically play a role in the situation, referred to as \textit{participants}. In the example, script knowledge subsumes the fact that the \textit{paying} event is a part of the \textit{eating in a restaurant} scenario.

Recent years have seen different approaches to learning script knowledge, centered around two strands: Work around narrative chains that are learned from large text collections \cite{chambers2008unsupervised,chambers2009unsupervised}, and the manual induction of script knowledge via crowdsourcing \cite{regneri2010learning,Wanzare2016}. Script knowledge has been represented both symbolically \cite{Jans2012,Pichotta2014,Rudinger2015} and with neural models \cite{modi:CONLL2014,Pichotta2016}. Scripts have been evaluated mostly intrinsically \cite{Wanzare2017,Ostermann2017}. An exception is \textit{MCScript} \cite{MCScript}, a reading comprehension corpus with a focus on script knowledge, and a predecessor to the data set presented in this work. Previous work has shown, however, that script knowledge is not required for performing well on the data set \cite{ostermann2018semeval}. 
Hence, to date, there exists no evaluation method that allows one to systematically assess the contribution of models of script knowledge to the task of automated text understanding. 

Our work closes this gap: We present \textit{MCScript2.0}, a reading \textbf{co}mprehension \textbf{co}rpus focused on \textbf{s}cript events and participants. It contains more than 3,400 texts about everyday scenarios, together with more than 19,000 multiple-choice questions on these texts. All data were collected via crowdsourcing. About half of the questions require the use of commonsense and script knowledge for finding the correct answer (like the question in Example \ref{ex:conversation}), a notably higher number than in MCScript. We show that in comparison to MCScript, commonsense-based questions in MCScript2.0 are also harder to answer, even for a model that makes use of a commonsense database. Thus, we argue that MCScript2.0 is the first resource which makes it possible to evaluate how far models are able to exploit script knowledge for automated text understanding. 

\begin{figure}
	\begin{tabularx}{\columnwidth}{|p{0.5cm}X|}
		\hline
		\textbf{T} & 
		(...) We put our ingredients together to make sure they were at the right temperature, preheated the oven, and pulled out the proper utensils. We then prepared the batter using eggs and some other materials we purchased and then poured them into a pan. After baking the cake in the oven for the time the recipe told us to, we then double checked to make sure it was done by pushing a knife into the center. We saw some crumbs sticking to the knife when we pulled it out so we knew it was ready to eat !
		\\
		\textbf{Q1} & \textit{When did they put the pan in the oven and bake it according to the instructions?}\\
		& After eating the cake. \ding{55}\\
		& After mixing the batter. \checkmark\\
		\textbf{Q2} & \textit{What did they put in the oven?}\\
		& The cake mix. \checkmark \\
		& Utensils. \ding{55}\\
		\hline
	\end{tabularx}
	\caption{Example text fragment from MCScript2.0}
	\label{fig:exampletext}
\end{figure}

Figure \ref{fig:exampletext} shows a text snippet from a text in MCScript2.0, together with two questions with answer alternatives\footnote{More text samples are given in the Supplemental Material.}. To find an answer for question 1, information about the temporal order of the steps for baking a cake is required: The cake is put in the oven after mixing the batter, and not after eating it---a piece of information not given in the text, since the event of putting the cake in the oven is not explicitly mentioned. Similarly, one needs script knowledge about which participants are typically involved in which events to know that the cake mix rather than the utensils is put into the oven. Both incorrect answer candidates are distractive: The utensils as well as the action of eating the cake are mentioned in the text, but wrong answers to the question.
Our contributions are as follows:
\begin{itemize}
	\item We present a new collecting method for challenging questions whose answers require commonsense knowledge and in particular script knowledge, as well as a new resource that was created with this method.
	\item We show that the task is simple for humans, but that existing benchmark models, including a top-scoring machine comprehension model that utilizes a resource for commonsense knowledge, struggle on the questions in MCScript2.0; especially on questions that require commonsense knowledge. 
	\item We compare MCScript2.0 to MCScript, the first machine comprehension resource for evaluating models of script knowledge. We show that in comparison to MCScript, the number of questions that require script knowledge is increased by a large margin and that such questions are hard to answer. Consequently, we argue that our dataset provides a more robust basis for future research on text understanding models that use script knowledge.
\end{itemize}

\section{Why another Machine Comprehension Dataset on Script Knowledge?}
\label{sec:motivation}
MCScript \cite{MCScript} is the first machine comprehension dataset designed to evaluate script knowledge in an end-to-end machine comprehension application, and to our knowledge the only other existing extrinsic evaluation dataset for script knowledge. Recent research has shown, however, that commonsense knowledge is not required for good performance on the dataset \cite{ostermann2018semeval,mitre}. 

We argue that this is due to the way in which questions were collected. During the collection process, workers were not shown a text, but only a very short description of the text scenario. As a result, many questions ask about general aspects of the scenario, without referring to actual details. This leads to the problem that there are many questions with standardized answers, i.e.~questions that can be answered irrespective of a concrete reading text. Examples \ref{ex:contribution6} and \ref{ex:contribution9} show two such cases, where the correct answer is almost exclusively \textit{in the shower} and \textit{on the stove}, independent of the text or even scenario.

\begin{exe}
	\ex Where did they wash their hair?\label{ex:contribution6}
	\ex Where did they make the scrambled eggs? \label{ex:contribution9}
\end{exe}

\citet{mitre} found that such information can essentially be learned from only the training data, using a simple logistic regression classifier and surface features regarding words in the text, question and answer candidates.

Also, many questions require vague inference over general commonsense knowledge rather than script knowledge. Example \ref{ex:contribution8} illustrates this: The simple fact that planting a tree gets easier if you have help is not subsumed by script knowledge about planting a tree, but a more general type of commonsense knowledge.

\begin{exe}
	\ex \textit{Text}: Once you know where to dig , select what type of tree you want. (...) Dig a hole large enough for the tree and roots . Place the tree in the hole and then fill the hole back up with dirt . (...)
	\\\textbf{Q}: Would it have been easier to plant the tree if they had help?
	\\ yes \checkmark
	\\ no \ding{55}
	\label{ex:contribution8}
\end{exe}

We inspected a random sample of 50 questions from the publicly available development set that were misclassified by the logistic model of \citet{mitre}. We found that for over 90\% of the inspected questions, the use of script knowledge would be only marginally relevant.

\paragraph{} 

We present a new data collection method and corpus that results in a larger number of challenging questions that require script knowledge. In particular, we define a revised question collection procedure, which ensures a  large proportion of non-trivial commonsense questions. 

\section{Corpus Creation}
\begin{figure*}[ht]
	\centering
	\fbox{\includegraphics[width=\textwidth]{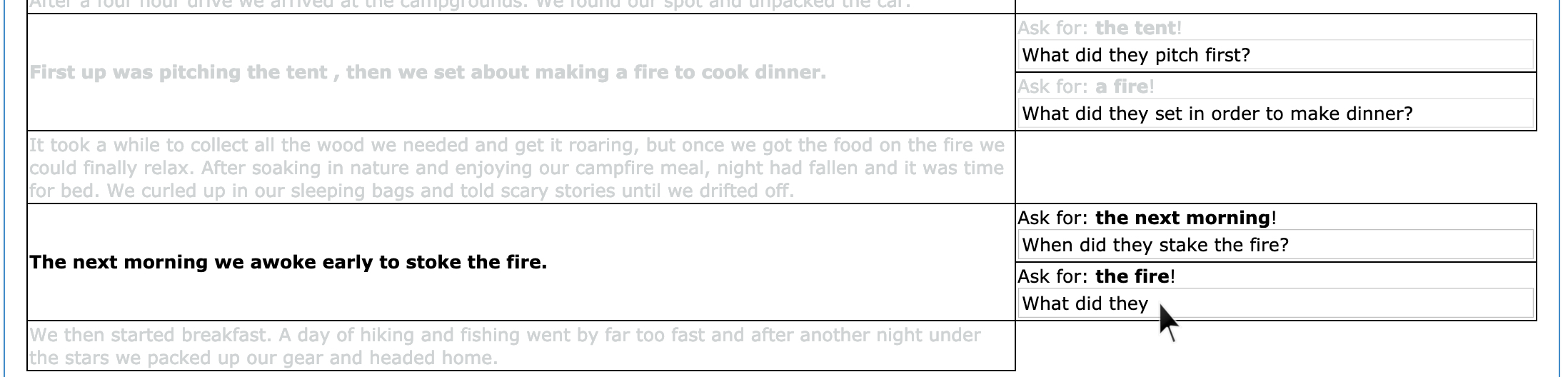}}
	\caption{Screenshot of an item in the participant question collection experiment.}
	\label{fig:np-experiment}
	
\end{figure*}

Texts, questions, and answer candidates are required for a multiple choice machine comprehension dataset. Our data collection process for texts and answers is based on the MCScript data and the methods developed there, but with several crucial differences. Like \citet{MCScript}, we create the data set via crowdsourcing. The question collection is revised to account for the shortcomings found with MCScript.

Similarly to \citet{MCScript}, we are interested in questions that require inference over script knowledge for finding a correct answer. Creating such questions is challenging: When questions are collected by showing a reading text and asking crowdsourcing workers to write questions, their answer can usually be read off the text. The authors of MCScript thus decided to not show a reading text at all, but only a short summary of the text scenario. This resulted in too general questions, so we decided for a third option: We identified a number of \textit{target sentences} in the reading text and guided workers to formulate questions about script-related details in these sentences. The target sentences were then hidden from the text, meaning that relevant information would have to be inferred from common sense during the answer collection and also in the task itself.
In the following sections, we describe the three data collection steps in detail.

\subsection{Text Collection}
As a starting point, we reused all texts from MCScript (2,119 texts on 110 scenarios) for our data set. To increase the topical coverage and diversity of the data set, we added texts for 90 new scenarios to our collection. As for MCScript, we selected topically different and plausible everyday scenarios of varying complexity, which were not too fine-grained (such as \textit{opening a window}). The scenarios were taken from 3 sources: First, we extracted scenarios from several script collections \cite{Wanzare2016,regneri2010learning,Singh2002} that are not part of MCScript. Second, we inspected the \textit{spinn3r} blog story corpus \cite{burton2009icwsm}, a large corpus of narrative blog stories 
and identified additional scenarios in these stories. Third, we added new scenarios that are related to existing ones or that extend them.

We collected 20 texts per new scenario, using the same text collection method as \citet{MCScript}: We asked workers to tell a story about a certain everyday scenario ``as if talking to a child''. This instruction ensures that the resulting stories are simple in language and clearly structured. Texts collected this way have been found to explicitly mention many script events and participants \cite{modiinscript,MCScript}. They are thus ideal to evaluate script-based inference.

\subsection{Question Collection}
For the question collection, we followed \citet{MCScript} in telling workers that the data are collected for a reading comprehension task for children, in order to get linguistically simple and explicit questions. However, as mentioned above, we guide workers towards asking questions about target sentences rather than a complete text. 

As target sentences, we selected every fourth sentence in a text. In order to avoid selecting target sentences with too much or too little content, we only considered sentences with less than 20 tokens, but that contained 2 or more noun phrases.\footnote{All parameters were selected empirically, by testing different values and analyzing samples of the resulting data.}

In a series of pilot studies, we then showed the texts with highlighted target sentences to workers and asked them to write questions about these sentences. We however found, that in many cases, the written questions were too general or nonsensical. 

We concluded that an even more structured task was required and decided to concentrate on questions of two types: (1) questions that ask about participants, and (2) questions about the temporal event structure of a scenario. Participants are usually instantiated by noun phrases (NPs), while events are described by verb phrases (VPs). We thus used \textit{Stanford CoreNLP} \cite{manning2014stanford} to extract both NPs and VPs in the target sentences and split up the experiment into two parts: 
In the first part, workers were required to write questions that \textit{ask about the given noun phrase}. Figure \ref{fig:np-experiment} shows a screenshot of an item from the first part. The first column shows the reading text with the target sentence highlighted. The second columns shows all extracted phrases with a field for one question per phrase.\footnote{If the noun phrase was part of a prepositional phrase or a construction of the form ``NP \textit{of} NP'', we took the whole phrase instead, because it is more natural to ask for the complete phrase. In order to avoid redundancy, we only looked at NPs that had no other NPs as parents. We also excluded noun phrases that referred to the narrator (\textit{I}, \textit{me} etc.).} Full details of the experiment instructions are given in the Supplemental Material.

In the second part, we then asked workers to write a temporal question (when, how long, etc.) \textit{using the given verb phrase}. We found that an exact repetition of the NP instructions for the second part (``ask about the given verb phrase'') resulted in unnatural questions, so we adapted the instructions. A screenshot of the VP experiment is given in the Supplemental Material.

We showed each text to two workers and asked them to write one question per VP or NP. Workers were only allowed to work on either the VP or the NP part, since the instructions could easily be confused. In order to exclude yes/no questions, we did not accept inputs starting with an auxiliary or modal verb. Also, all questions needed to contain at least 4 words. We asked workers to use \textit{they} to refer to the protagonist of the story and other types of mentions (e.g.~pronouns like \textit{I}, \textit{you}, \textit{we} or the word \textit{narrator}) were not accepted.

\begin{table*}[t]
	\centering
	\begin{tabular}{c|c|c||c|c}
		\toprule
		\textit{text-based} & \textit{script-based} & \textit{text-or-script} & \textit{unfitting} & \textit{unknown} \\
		\hline
		9,357 & 12,433 & 2,403 & 3,240 & 6,457 \\
		\midrule
		\multicolumn{3}{c}{\textbf{total answerable:} 24,193} & \multicolumn{2}{c}{\textbf{total not answerable:} 9,397}\\
		\bottomrule
	\end{tabular}
	\caption{Distribution of question labels, before validation.} 
	\label{tab:results}
\end{table*}
\subsection{Answer Collection}

For collecting answer candidates we hid the target sentences from the texts and showed them with up to 12 questions, to keep the workload at an acceptable level. If there were more questions for a text, we selected 12 questions at random. 

Since the target sentences are hidden in the texts, it can be expected that some questions cannot be answered from the text anymore. However, the necessary information for finding an answer might be inferred from script knowledge, so workers were explicitly told that they might need commonsense to find an answer. Some answers can still be read off the text, if other parts of the texts contain the same information as the hidden target sentences. For other questions, neither the text nor script knowledge provides sufficient information for finding an answer.

As for the creation of MCScript, workers first had to conduct a 4-way classification for each question to account for these cases: \textit{text-based} (answer is in the text), \textit{script-based} (answer can be inferred from script knowledge), \textit{unfitting} (question doesn't make sense), \textit{unknown} (answer is not known). Having such class annotations is not only useful for evaluation, but it also sensitizes workers for the fact that they are explicitly allowed to use background knowledge.

In the experiment, workers were also instructed to write both a correct and a plausible incorrect answer for questions labeled as \textit{text-based} of \textit{script-based}. We follow \cite{MCScript} and require workers to write an alternative question if the labels \textit{unfitting} or \textit{unknown} are used, in order to level out the workload.

We presented each question to 5 workers, resulting in 5 judgements and up to 5 incorrect and correct answer candidates per question. For the final data set, we considered questions with a majority vote (3 out of 5) on \textit{text-based} or \textit{script-based}. We also included questions without a majority vote, but for which at least 3 workers assigned one of \textit{text-based} or \textit{script-based}. In that case, we assigned the new label \textit{text-or-script} and also accepted the question for the final data set. This seemed reasonable, since at least 3 workers wrote answers for the question, meaning it could still be used in the final data collection. The remaining questions were discarded. 

\subsection{Answer Candidate Selection}
\label{sec:answer-selection}
In a last step, we selected one correct and one incorrect answer from all possible candidates per question for the data set. To choose the most plausible correct answer candidate, we adapt the procedure from \citet{MCScript}: We normalize all correct answers (lowercasing, normalizing numbers\footnote{We used \textit{text2num}, \url{https://github.com/ghewgill/text2num}.}, deleting stopwords\footnote{\textit{and}, \textit{or}, \textit{to}, \textit{the}, \textit{a}}) and then merge candidates that are contained in another candidate, and candidate pairs with a \citet{levenshtein1966binary} distance of less than 3. The most frequent candidate is then selected as correct answer. If there was no clear majority, we selected a candidate at random.

\par To select an incorrect answer candidate, we adapt the \textit{adversarial filtering} algorithm from \citet{zellers2018swag}. Our implementation uses a simple classifier that utilizes shallow surface features. The algorithm selects the incorrect answer candidate from the set of possible candidates that is most difficult for the classifier, i.e.~an incorrect answer that is hard to tell apart from the correct answer (e.g.~the incorrect answers in Figure \ref{fig:exampletext}: \textit{eating} and \textit{utensils} are also mentioned in the text). By picking incorrect answers with the adversarial filtering method, the dataset becomes robust against surface-oriented methods.

\par Practically, the algorithm starts with a random assignment, i.e.~a random incorrect answer candidate per question. This assignment is refined iteratively, such that the most difficult candidate is selected. In each iteration, the algorithm splits the data into a random training part and a test part. The classifier is trained on the training part and then used to classify \textit{all} possible candidates in the test part. The assignment of answer candidates in the test data is then changed such that the most difficult incorrect answer candidate per question is picked as incorrect answer. After several iterations through the whole dataset, the number of changed answer candidates usually stagnates and the algorithm converges.

For MCScript2.0, we use the logistic classifier mentioned in Section \ref{sec:motivation}, which only uses surface features and is thus well suited for the filtering algorithm. Implementation details and pseudocode are given in the Supplemental Material.

\section{Corpus Analysis}
\subsection{General Statistics}
\begin{figure*}[t]
	\centering
	\includegraphics[width=.5\textwidth]{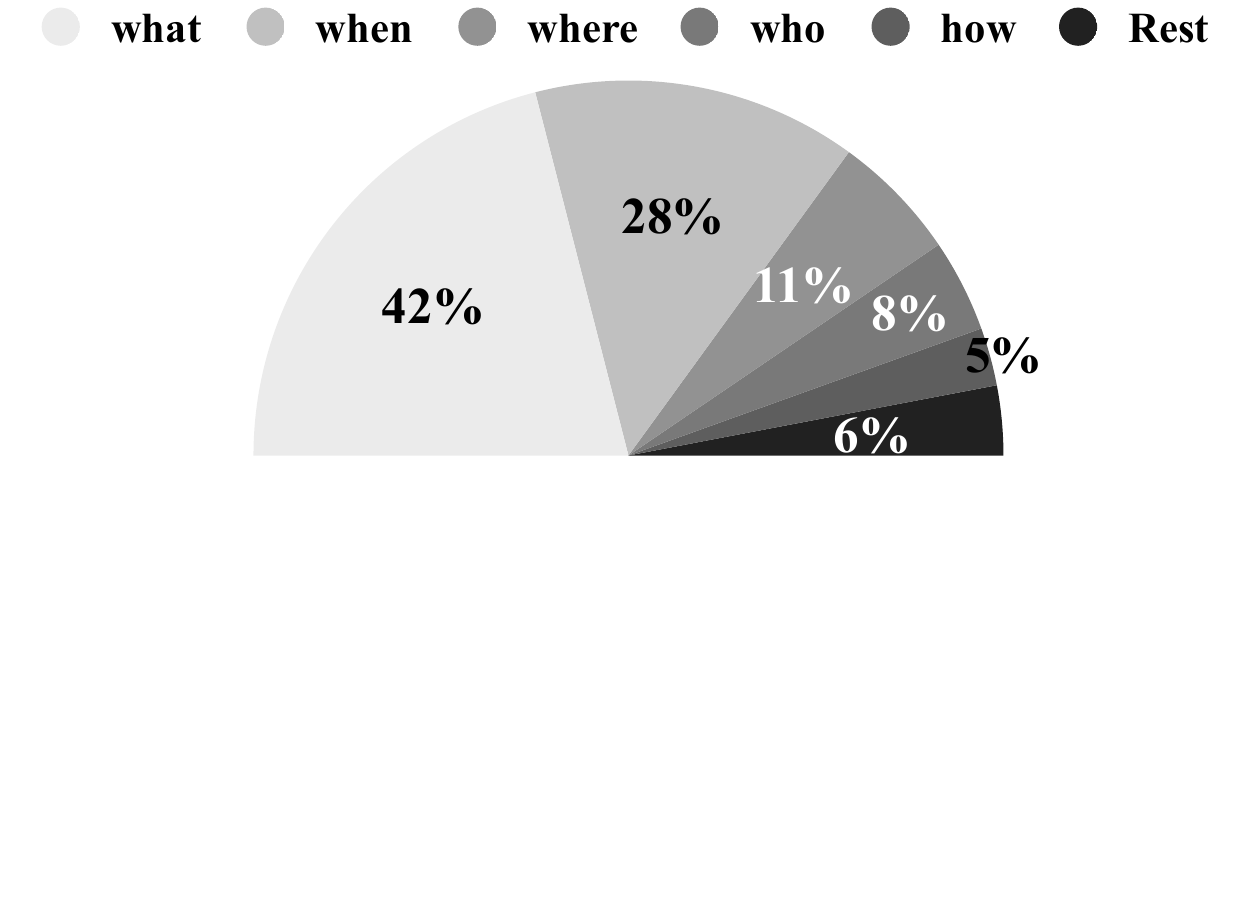}
	\vspace{-3cm}
	\caption{Distribution of question types}
	\label{fig:questiontypes}
\end{figure*}
\begin{figure*}[t]
	\centering
	\includegraphics[width=.67\textwidth]{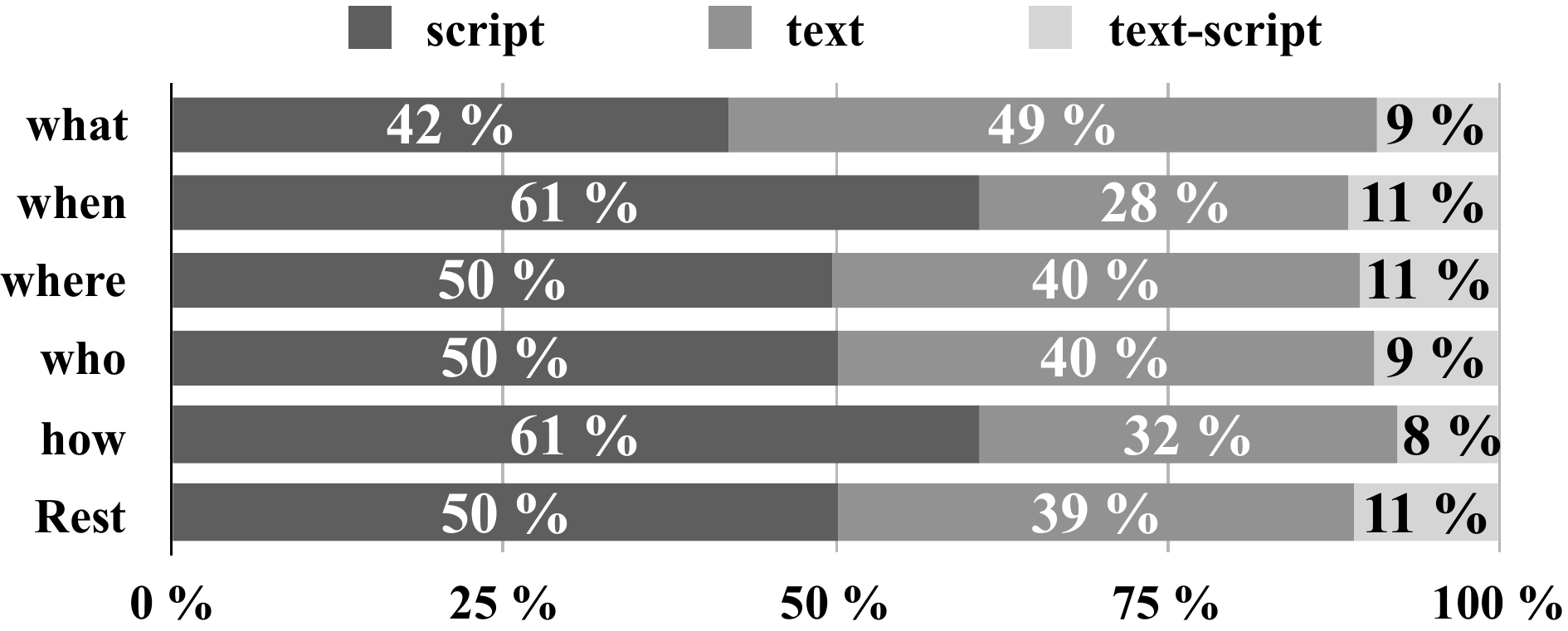}
	\caption{Proportion of labels per question type.}
	\label{fig:typesXlabels}
\end{figure*}

In total, MCScript2.0 comprises 19,821 questions on 3,487 texts, i.e.~5.7 questions on average per text. The average length of texts, questions and answers is 164.4 tokens, 8.2 tokens and 3.4 tokens, respectively.

In the data collection process, we crowdsourced 1,800 new texts, resulting in a total of 3,919 texts for 200 scenarios. On average, there are 1.98 target sentences per text. 
In the question collection, we gathered 42,132 questions that were then used for the answer collection.
For 8,242 questions, there was no clear majority on the question label. Table \ref{tab:results} shows the label distribution on the remaining 33,890 questions. 24,193 of these could be answered, i.e.~71\%. 

To increase data quality, we conducted a manual validation of the data. Four student assistants replaced erroneous answers and deleted nonsensical questions, question duplicates and incoherent texts. 

During validation, 152 texts were found to be incoherent and discarded (along with all questions). Additionally, 3,388 questions were deleted because they were nonsensical or duplicates. 1,620 correct and 2,977 incorrect answers were exchanged, resp., because they were inappropriate. If a question deletion resulted in texts without any questions, or if a text did not have any answerable questions, the text was discarded, too.

After question validation, the final dataset comprises 9,935 questions that are labeled as script-based, 7,908 as text-based, and 1,978 as text-or-script. 

\subsection{Questions} 
Figure \ref{fig:questiontypes} gives the distribution over question types, which we extracted by looking at the first word in the question. The largest number of questions are \textit{what} questions, most of which ask about participants of a script. \textit{When} questions make up the second largest group, asking for temporal event structure. During the VP question experiment, some workers ignored that we asked for temporal questions only, which resulted in a number of \textit{how} questions. 

MCScript2.0 contains 50\% questions labeled as script-based, which is a notably larger amount as compared to the approximately 27\% of questions in MCScript labeled as script-based. The number of script-based questions varies between the question types, as can be seen in Figure \ref{fig:typesXlabels}. While \textit{when} and \textit{how} questions require script knowledge for finding an answer in more than 60\% of cases, less than half of \textit{what} questions do so. A simple explanation for this could be that \textit{when} or \textit{how} questions typically ask for events, while \textit{what} questions ask for participants. Events are usually referred only once in a text, i.e.~with the hiding of the respective event mention, the needed information has to be inferred. Participants in contrast tend to appear more often throughout a story.

Example \ref{ex:10} below illustrates this. Question 1 was originally asked about a sentence in which the plates are set for the dinner guests. The guests still appear in another sentence, so the answer can be inferred from the text. 

For question 2, in contrast, script knowledge is required for finding an answer: The event of \textit{bringing the items to the table} is not mentioned anymore, so the information that this happens typically after counting plates and silverware needs to be inferred.

\begin{exe}
	\ex \textbf{T:}
	(...) I was told that there would be 13 or 14 guests. First I counted out 14 spoons, then the same number of salad forks, dinner forks, and knives. (...) 
	I set each place with one napkin, one dinner fork, one salad fork, one spoon, and one knife. (...)\\
	\textbf{Q1:} Who are the plate and cup for?\\
	dinner guests \checkmark ~~~~~ the neighbor \ding{55}\\
	\textbf{Q2:} When did they bring the items over to the table?\\
	after counting them \checkmark\\
	after placing them on the table \ding{55}\\
	\label{ex:10}
\end{exe}

\section{Experiments}

We test three benchmark models on MCScript2.0 that were also evaluated on MCScript, so a direct comparison is possible. For the experiments, we split the data into a training set (14,191 questions on 2,500 texts), a development set (2,020 questions on 355 texts) and a test set (3,610 questions on 632 texts). All texts of 5 randomly chosen scenarios were assigned completely to the test set, so a part of the test scenarios are unseen during training.

\subsection{Models}

\subsubsection*{Logistic Regression Classifier}
As first model, we reimplemented the logistic regression classifier proposed by \citet{mitre}, which was also used in the adversarial filtering algorithm. The classifier employs 3 types of features: (1) Length features, encoding the length of the text, answer and questions on the word and character level, (2) overlap features, encoding the amount of literal overlap between text, question, and answers, and (3) binary lexical patterns encoding the presence or absence of words or combinations of words in answer, text and question.

\subsubsection*{Attentive Reader}
As second model, we implement an attentive reader \cite{hermann2015teaching}. 
We adopt the formulation by \citet{MCScript}  (originally by \citet{chen2016thorough}). All tokens in text, question and answers are represented with word embeddings. Bi-directional gated recurrent units (GRUs, \citet{cho2014learning}) process the text, question and answers and transform them into sequences of contextualized hidden states. The text is represented as a weighted average of the hidden states with a bilinear attention formulation, and another bilinear weight matrix is used to compute a scalar as score for each answer. For a formalization, we refer to \citet{MCScript} and \citet{chen2016thorough}. 

\subsubsection*{Three-way Attentive Network (TriAN)}
As third model, we use a three-way attentive network \citep{yuanfudao}, the best-scoring model of the shared task on MCScript\footnote{Code available at \url{https://github.com/intfloat/commonsense-rc}}. Various types of information are employed to represent tokens: Word embeddings, part of speech tags, named entity embeddings, and word count/overlap features, similar to the logistic classifier. Three bidirectional LSTM \citep{hochreiter1997long} modules are used to encode text, question and answers. The resulting hidden representations are reweighted with three attention matrices and then summed into vectors using three self-attention layers. 

Additionally, token representations are enhanced with \textit{ConceptNet} \cite{speer2017conceptnet} relations as a form of induced commonsense knowledge. ConceptNet is a large database of commonsense facts, represented as triples of two entities with a predicate. Relevant ConceptNet relations between words in the answer and the text are queried from the database and represented with relation embeddings, which are learned end-to-end during training and appended to the text token representations.

In contrast to \citet{yuanfudao}, we use the non-ensemble version of TriAN without pretraining on \textit{RACE} \cite{lai2017race}, for better comparability to the other models.

\subsection{Human Upper Bound}
To assess human performance, 5 student assistants performed the reading comprehension task on 60 texts each. To assess agreement, 20 texts were annotated by all students. The annotators reached averaged pairwise agreement of 96.3\% and an average accuracy of 97.4\%, which shows that this is a simple task for humans. 
\begin{figure*}
	\centering
	\includegraphics[width=.8\textwidth]{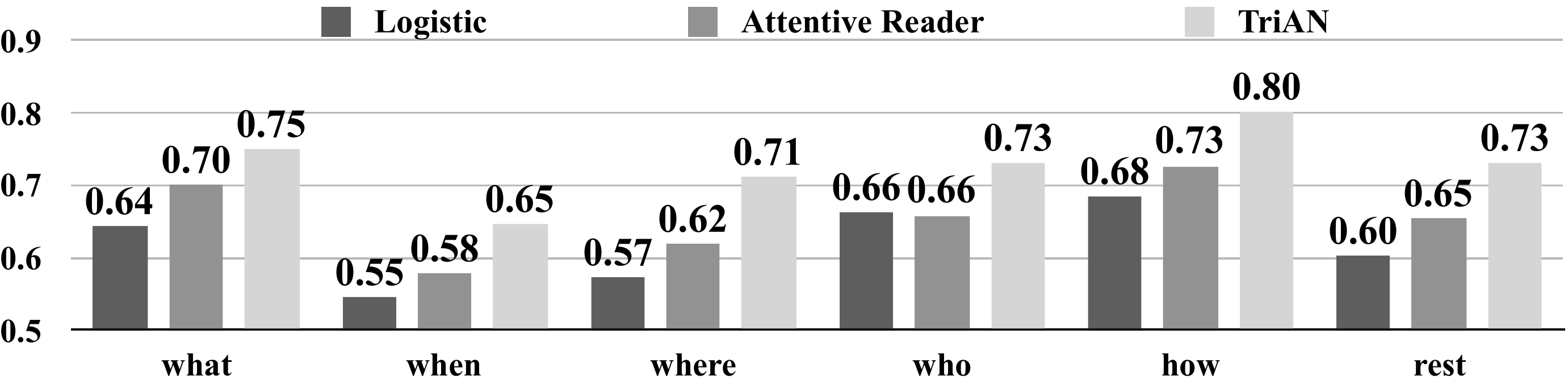}
	\caption{Performance of the models on question types.}
	\label{fig:performance}
\end{figure*}
\subsection{Results}

\textbf{Overall Performance.} Table \ref{tab:performance} gives details about the performance of the 3 benchmark models on the test set, and on script-based (acc$_{scr}$) and text-based (acc$_{txt}$) questions in the test set. As can be seen, the logistic model scores worst, presumably because it has been used for the adversarial filtering algorithm and the data are thus most challenging for this model. 
TriAN performs best, clearly outperforming the attentive reader. TriAN is apparently superior in its way of text processing, since it employs a richer text representation and exploits attention mechanisms on more levels, which is reflected by a higher accuracy on text-based questions. In contrast, script-based questions seem to be challenging for TriAN. This is interesting, because it shows that ConceptNet alone cannot provide sufficient information for answering the kind of questions that can be found in MCScript2.0.

\textbf{Comparison to MCScript.} Since the same models were used for MCScript, a comparison of their performance on both datasets is possible. Results on MCScript are given in Table \ref{tab:performance-mcscript}.\footnote{For the attentive reader, numbers were taken from \cite{ostermann2018semeval}. The other models were retrained (and in the case of the logistic model re-implemented), since no exact numbers on script/text-based questions were published.}
As can be seen, the performance of all three models is worse on MCScript2.0, showing that the dataset is generally more challenging. In contrast to MCScript, script-based questions in MCScript2.0 are clearly harder to answer than text-based questions: All models perform worse on script-based questions as compared to text-based questions. In comparison to MCScript, the performance of TriAN is \textbf{12\%} lower. This indicates that the new mode of question collection and the answer selection via adversarial filtering resolve some of the difficulties with MCScript. 

\begin{table}[t]
	\begin{tabularx}{\columnwidth}{p{2.6cm}XXX}
		\toprule
		& \textbf{~acc}  & \textbf{acc$_{scr}$} & \textbf{acc$_{txt}$}\\
		\midrule
		\textit{Logistic Model}    &0.61 & 0.56 & 0.67\\
		\textit{Attentive Reader}  & 0.65 & 0.63 & 0.68\\
		\textit{TriAN}   & \textbf{0.72} & \textbf{0.67} & \textbf{0.78}\\ 
		\midrule
		\midrule
		\textit{Humans} &  \multicolumn{3}{l}{0.97}\\
		\bottomrule
	\end{tabularx}
	\caption{Accuracy on test set, and on script/text-based questions (acc$_{scr}$, acc$_{txt}$) on \textbf{MCScript2.0}. The maximum per column is printed in \textbf{bold}.}
	\label{tab:performance}
\end{table}

To assess whether the performance difference to MCScript is due to the 90 new scenarios being more challenging, we additionally evaluated the models on these scenarios only. We found no performance difference on the new vs.~old scenarios.

\textbf{Influence of Adversarial Filtering.} To find out how large the influence of the new question collection method and the answer selection via adversarial filtering is, we conducted an additional experiment: We applied the answer selection method of \citet{MCScript} to our data set to create an alternative version of the data that is not based on adversarial filtering. Correct answers were selected to have the lowest possible overlap with the reading text. Incorrect answers were selected using the majority voting technique described in Section \ref{sec:answer-selection}. 

We found that the adversarial filtering accounts for around two thirds of the total accuracy difference of TriAN as compared to MCScript, i.e.~one third of the difference can be attributed to the new question collection. This means that both modifications together add to the larger difficulty of MCScript2.0. 
\begin{table}[t]
	\begin{tabularx}{\columnwidth}{p{2.6cm}XXX}
		\toprule
		&  \textbf{~acc}  & \textbf{acc$_{scr}$} & \textbf{acc$_{txt}$}\\
		\midrule
		\textit{Logistic Model}   & 0.79 & ~0.76 &  ~\textbf{0.81}\\
		\textit{Attentive Reader} & 0.72 & ~0.75 & ~0.71\\
		\textit{TriAN} & \textbf{0.80} & ~\textbf{0.79} & ~\textbf{0.81}\\ 
		\midrule
		\midrule
		\textit{Humans} &  \multicolumn{3}{l}{0.98}\\
		\bottomrule
	\end{tabularx}
	\caption{Accuracy on the test set and on script/text-based questions (acc$_{scr}$, acc$_{txt}$) on \textbf{MCScript}. The maximum per column is printed in \textbf{bold}.}
	\label{tab:performance-mcscript}
\end{table}

\textbf{Question Types.} Figure \ref{fig:performance} shows the performance of the models on single question types, as identified in Section 4. It is clear that \textit{when} questions are most challenging for all models. The logistic classifier performs almost at chance level. As far as TriAN is concerned, we found that many cases of errors ask for the typical temporal order of events, as Example \ref{ex:missc} illustrates:

\begin{exe}
	\ex \textbf{Q:} When did they put the nozzle in their tank? \\
	before filling up with gas. \checkmark \\
	after filling up with gas.  \ding{55}
	\label{ex:missc}
\end{exe}

The event of \textit{put the nozzle in the tank} is not mentioned in the shown version of the text, so it is not possible to read off the text when the event actually took place.

\textit{How} questions are the least difficult questions. This can be explained with the fact that many \textit{how} questions ask for numbers that are mentioned in the text (e.g.~\textit{How long did they stay in the sauna?} or \textit{How many slices did they place onto the paper plate?}). The answer to such questions can often be found with a simple text lookup. Another part of how questions asks for the typical duration of an activity. These questions often have similar answers irrespective of the scenario, since most of the narrations in MCScript2.0 span a rather short time period. Such answers can easily be memorized by the models.

Especially for TriAN, \textit{what} and \textit{who} questions seem to be easy. This could be explained with the fact that ConceptNet contains lots of information about entities and their relations to each other, apparently also covering some information about script participants, which seems to be useful for these question types.

\section{Related Work}
Recent years have seen a number of datasets that evaluate commonsense inference. Like our corpus, most of these data sets choose a machine comprehension setting. The data sets can be roughly classified along their text domain:

\textbf{News Texts.} Two recently published machine comprehension data sets that require commonsense inference are based on news texts. First, \textit{NewsQA} \citep{trischler2017newsqa} is a dataset of newswire texts from CNN with questions and answers written by crowdsourcing workers. During data collection, full texts were not shown to workers as a basis for question formulation, but only the text's title and a short summary, to avoid literal repetitions and support the generation of non-trivial questions requiring background knowledge. 
Second, \textit{ReCoRD} \cite{zhang2018record} contains cloze-style questions on newswire texts that were not crowdsourced, but automatically extracted by pruning a named entity in a larger passage from the text.

\textbf{Web Texts.} Other corpora use web documents. An example is \textit{TriviaQA} \citep{JoshiTriviaQA2017}, a corpus that contains automatically collected question-answer pairs from 14 trivia and quiz-league websites, together with web-crawled evidence documents from \textit{Wikipedia} and \textit{Bing}. While a majority of questions require world knowledge for finding the correct answer, it is mostly factual knowledge. 
\textit{CommonsenseQA} \citep{talmor2018commonsenseqa} contains a total of over 9000 multiple-choice questions that were crowdsourced based on knowledge base triples extracted from ConceptNet. Texts were only added post-hoc, by querying a web search engine based on the formulated question, and by adding the retrieved evidence texts to the questions and answers. 

\textbf{Fictional Texts.} \textit{NarrativeQA} \cite{kovcisky2018narrativeqa} is a reading comprehension dataset that largely differs from other corpora by means of text length. Instead of providing short reading texts, models have to process complete books or movie scripts and answer very complex questions. 

Because of their domains, the aforementioned data sets evaluate a very broad notion of commonsense, including e.g.~physical knowledge (for trivia texts) and knowledge about political facts (for newswire texts). However, none of them explicitly tackle script knowledge.

\section{Conclusion}
We presented MCScript2.0, a new machine comprehension dataset with a focus on challenging inference questions that require script knowledge or commonsense knowledge for finding the correct answer. Our new question collection procedure results in about half of the questions in MCScript2.0 requiring such inference, which is a much larger amount compared to a previous dataset. 

We evaluate several benchmark models on MCScript2.0 and show that even a model that makes use of ConceptNet as a source for commonsense knowledge struggles to answer many question in our corpus. MCScript2.0 forms the basis of a shared task organized at the COIN workshop.\footnote{\url{https://coinnlp.github.io/}}

\section*{Acknowledgments}

We thank our student assistants Leonie Harter, David Meier, Christine Sch\"afer and Georg Seiler for the help with data validation, and Kathryn Chapman, Srestha Ghosh, Trang Hoang, Ben Posner and Miriam Schulz for help with assessing the human upper bound. We also thank the numerous workers on MTurk for their good work and Carina Silberer and the reviewers for their helpful comments on the paper. This research was funded by the German Research Foundation (DFG) as part of SFB 1102 ‘Information Density and Linguistic Encoding’.

\bibliography{references}
\bibliographystyle{aclnatbib}

\appendix

\newpage
\section{Supplemental Material}
\subsection{Additional Data Sample}

\begin{exe}
	\ex \textbf{T:} I am at work . I have a guest sit at the bar . The ordered themselves a beer . I check that he is of age , and that his license is valid . I then go to the beer cooler , and grab a nice cold mug , and fill it up with beer . I place a napkin down and set the beer on top in front of the bar guest . I present him the check and tell him no rush , whenever he is ready . He then places his cash with the receipt . I go to cash him out , offer to be right back with his change , and he responds with , " Keep the change " . I like nights like this .\\
	\textbf{Q:} Why did they receive a nice tip?\\
	the customer was happy with the service \checkmark\\
	the customer was in a rush \ding{55}\\
	\textbf{Q:} When was the check printed?\\
	after the order \checkmark\\
	before the order \ding{55}\\
	\textbf{Q:} What did they create at the computer and print?\\
	the check \checkmark\\
	change \ding{55}\\
\end{exe}

\begin{exe}
	\ex \textbf{T:} I wanted to throw a Bachelorette Party for my best friend . She lives in Dallas , but she wants to have her party in New Orleans for a girls weekend . The first thing we did was talk about the theme of the party . We decided on the theme of `` Something Blue '' . We would have all colors of blue and activities that have titles with the word blue for the whole weekend . She gave me a list of 20 girls . I created an invitation that had blue and included a picture of her . I also included an itinerary of our weekend activities with all of our fun `` blue '' titles , to set the fun mood . I sealed them before hand writing the addresses and adding a stamp . Next , they were off to the post office , so everyone could be invited to our fun weekend .\\
	\textbf{Q:} What was printed out?\\
	itinerary \checkmark\\
	invitations to a weddingy \ding{55}\\
	\textbf{Q:} When was each invitation placed into their blue envelope?\\
	Before handwriting addresses \checkmark\\
	After adding stamps \ding{55}\\
	\textbf{Q:} Where did she place the invitations?\\
	Post office \checkmark\\
	Dallas \ding{55}\\
\end{exe}

\begin{exe}
	\ex \textbf{T:} The restaurant was terrible again and I probably should not have given it another chance . The management at the store level is obviously not paying attention to me so it is time to right to headquarters . I opened the word processing program on my computer and opened a new document .  I went all the way to the right side and entered my street address on one line and the city , state and zip code below that .  Next I entered the date and then moved all the way to the left and entered the street address of the restaurant headquarters and the city , state and zip code of the headquarters .  I started the letter with Dear Sir and on the next lines , proceeded to explain the problems I had been having with this particular location , it 's service and food . I explained that I had tried to resolve it at store level but had been unsuccessful . On the final line , I went all the way to right and entered ` Sincerely ' and hit return a couple times , then added my name below that .  I folded it and put it an addressed and stamped envelope , and mailed it to the company headquarters .\\
	\textbf{Q:} What did they print out?\\
	The letter  \checkmark\\
	The receipt from the restaurant \ding{55}\\
	\textbf{Q:} When did they sign above their printed name?\\
	After the letter was printed \checkmark\\
	After putting the letter in the envelope. \ding{55}\\
	\textbf{Q:} What they did they sign?\\
	The letter \checkmark\\
	The receipt at the restaurant \ding{55}\\
\end{exe}

\subsection{Crowdsourcing Details}
All data were collected using Amazon Mechanical Turk\footnote{\url{https://www.mturk.com}}. We paid \$0.50 for each item in the text and answer collection experiment. For the question collection experiment, we paid \$0.50 per item, if the text contained 4 or more target sentences, and \$0.30 per item if fewer target sentences were highlighted. 

\subsection{Implementation Details}
For implementation details and preprocessing of the logistic model and TriAN, we follow \cite{mitre} and \cite{yuanfudao}, respectively. NLTK\footnote{\url{https://www.nltk.org}} was used as preprocessing tool for the Attentive Reader.

The learning rate was tuned to 0.002 and 0.1 for TriAN and the attentive reader, resp. and the hidden size for both models to 64. As in the original formulation, dropout was set to 0.5 for the attentive reader and to 0.4 for TriAN. Batch size was set to 32 and both models were trained for 50 epochs. The model with the best accuracy on the dev data was used for testing. 

\subsection{Adversarial Filtering}
\begin{algorithm}
	\KwData{data set $D$, a randomly initialized assignment $S$, and a classifier $C$}
	\KwResult{$\hat{S}$}
	\Repeat{number of changed assignments stagnates or increases}{
		split the data into test batches of size $b$, such that each batch contains all questions for $b$ texts\;
		\For{$D_{test}$ in batches}{
			$D_{train} \longleftarrow D\backslash D_{test}$\;
			$\mathcal{D}_{train} \longleftarrow compile(D_{train})$\;
			train $C$ on $\mathcal{D}_{train}$\;
			\For{all instances $<T_i, Q_i, a_i^+, <a_{i,0}^-...a_{i,j}^->$ in $D_{test}$}{
				use $C$ to classify all incorrect answer candidates $a_{i,0}^-...a_{i,j}^-$\;
				set $s_{i,j}$ to the index of the answer candidate with the highest probability of being correct\;
			}
		}
	}
	\caption{Adversarial Filtering for MCScript2.0}
	\label{alg:1}
\end{algorithm}
Formally, let a dataset be defined as a list of tuples $\langle t_i, q_i, a_i^+, \langle a_{i,0}^-...a_{i,j}^-\rangle\rangle$, where $t_i$ is a reading text, $q_i$ is a question on the text, $a_i^+$ is the correct answer (as selected via majority vote, s. last Section) and  $\langle a_{i,0}^-...a_{i,j}^-\rangle$ is a list of 3 to 5 incorrect answer candidates\footnote{Note that since there are several questions per text, the value of $t_i$ may appear in several instances.}. The aim of the algorithm is to find an assignment $\hat{S} = \{s_{0,0} ... s_{i,j}\}$, where each $s_{i,j}$ is the index of the most difficult answer candidate in $\langle a_{i,0}^-...a_{i,j}^-\rangle$. 

A dataset that is \textit{compiled} with the assignment $S$ is a list of instances $<t_i, q_i, a_i^+, a_i^->$, such that there is only one incorrect answer candidate per question, according to the indices given by $S$.

Once the algorithm converges, $\hat{S}$ is used to compile the final version of the dataset, $\hat{\mathcal{D}}$, which contains incorrect answer candidates that are most likely to be correct.

For the batch size we tried values in $\{50, 100, 250, 500\}$, but we found that for all values, the performance of the classifier would drop close to chance level after one iteration only. We set $b = 250$, since the performance was closest to chance after convergence with that setting. Also, we defined that the algorithm converges if the number of changed assignments since the last iteration is $\leq 50$. 

\ \\\ \\\ \\\ \\\ \\\ \\\ \\\ \\\ \\\ \\\ \\\ \\\ \\\ \\\ \\\ \\\ \\\ \\\ \\\ \\\ \\\ \\\ \\\ \\\ \\
\subsection{Screenshot of the VP-based Question Collection Experiment.}
\begin{figure}[h]
	\centering
	\includegraphics[width=\textwidth]{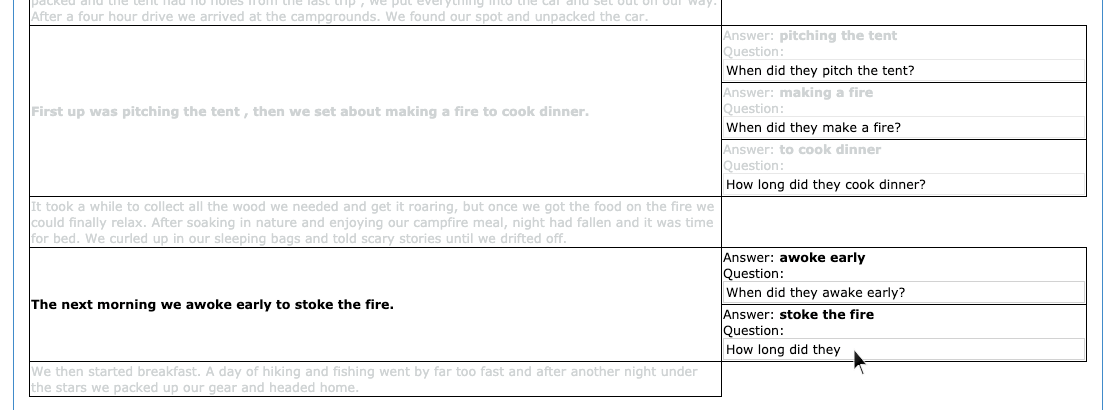}
\end{figure}
\ \\\ \\\ \\\ \\\ \\\ \\\ \\\ \\\ \\\ \\\ \\\ \\\ \\\ \\\ \\\ \\\ \\\ \\\ \\\ \\\ \\\ \\\ \\\ \\\ \\\ \\\ \\\ \\\ \\\ \\\ \\\ \\\ \\\ \\\ \\\ \\\ \\\ \\\ \\\ \\\ \\\ \\\ \\\ \\\ \\\ \\\ \\\ \\\ \\\ \\\ \\\ \\\ \\\ \\\ \\\ \\\ \\\ \\\ \\\ \\\ \\\ \\\ \\\ \\\ \\\ \\\ \\\ \\\ \\\ \\\ \\\ \\\ \\\ \\\ \\\ \\\ \\\ \\\ \\\ \\\ \\\ \\\ \\\ \\\ \\
\subsection{Screenshot of the Instructions for the NP-based Question Collection Experiment.}
\begin{figure}[h]
	\centering
	\includegraphics[width=\textwidth]{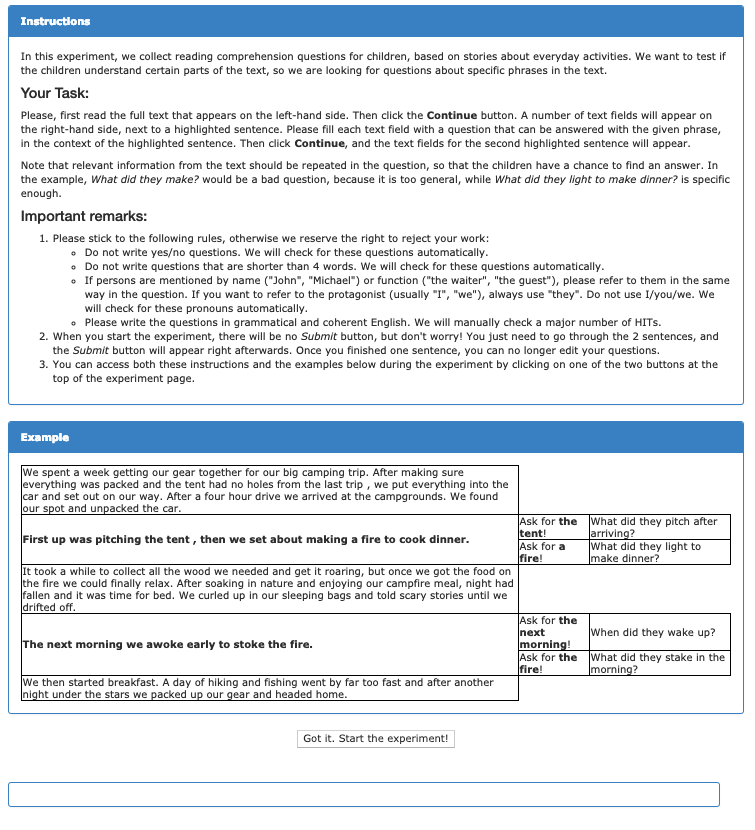}
\end{figure}

\end{document}